%% file: main.tex
\documentclass[10pt,twocolumn,letterpaper]{article}

\PassOptionsToPackage{numbers,sort&compress}{natbib}
\PassOptionsToPackage{hyphens}{url}
\PassOptionsToPackage{pagebackref,breaklinks,colorlinks,allcolors=cvprblue}{hyperref}
\PassOptionsToPackage{dvipsnames,table}{xcolor}

\usepackage{microtype}
\usepackage{graphicx}
\usepackage{subfigure}
\usepackage{booktabs} 

\usepackage[dvipsnames, table]{xcolor}

\usepackage{amsmath}
\usepackage{amssymb}
\usepackage{mathtools}
\usepackage{amsthm}

\usepackage{booktabs} 
\usepackage{graphicx}
\usepackage{amssymb}
\usepackage{caption}
\usepackage{subcaption}
\usepackage{multirow}
\usepackage{overpic}
\usepackage{textpos}
\usepackage[british, english, american]{babel}

\definecolor{citecolor}{HTML}{0071BC}
\definecolor{linkcolor}{HTML}{ED1C24}

\newlength\savewidth\newcommand\shline{\noalign{\global\savewidth\arrayrulewidth
  \global\arrayrulewidth 1pt}\hline\noalign{\global\arrayrulewidth\savewidth}}

\renewcommand{\paragraph}[1]{\vspace{1.25mm}\noindent\textbf{#1}}

\newcolumntype{x}[1]{>{\centering\arraybackslash}p{#1pt}}
\newcolumntype{y}[1]{>{\raggedright\arraybackslash}p{#1pt}}
\newcolumntype{z}[1]{>{\raggedleft\arraybackslash}p{#1pt}}

\newcommand{\app}{\raise.17ex\hbox{$\scriptstyle\sim$}}

\definecolor{deemph}{gray}{0.6}

\definecolor{baselinecolor}{gray}{.9}

\usepackage{natbib}

\usepackage{bbding}

\theoremstyle{plain}

\theoremstyle{definition}

\theoremstyle{remark}

\usepackage[textsize=tiny]{todonotes}

\usepackage{hyperref}
\usepackage{cvpr}              


\definecolor{cvprblue}{rgb}{0.21,0.49,0.74}
\usepackage[pagebackref,breaklinks,colorlinks,allcolors=cvprblue]{hyperref}

\def\paperID{38} 
\def\confName{CVPR}
\def\confYear{2025}

\title{Prototype Augmented Hypernetworks for Continual Learning}

\author{
    Neil De La Fuente\textsuperscript{1,2}\thanks{Equal contribution. Correspondence: \texttt{neil.de@tum.de}} \quad 
    Maria Pilligua\textsuperscript{1}\footnotemark[1] \quad  
    Daniel Vidal\textsuperscript{1,2} \quad \\
    Albin Soutiff\textsuperscript{} \quad
    Cecilia Curreli\textsuperscript{2,3} \quad
    Daniel Cremers \textsuperscript{2,3} \quad
    Andrey Barsky\textsuperscript{1}
    \\[1.5mm] 
    \textsuperscript{1}Computer Vision Center \qquad 
    \textsuperscript{2}TUM \qquad
    \textsuperscript{3}Munich Center for Machine Learning \\ %
}

\begin{document}
\maketitle

\input{sec/0_abstract}

\vspace{-0.2cm}
\section{Introduction}

\label{sec:introduction}

Continual learning (CL) enables models to learn sequential tasks without forgetting prior knowledge, but catastrophic forgetting (CF) remains a major obstacle. Addressing CF is crucial for developing AI systems that can adapt to evolving environments, such as in robotics, autonomous driving, or personalized healthcare~\cite{verwimp2023continual}. Recent work suggests CF primarily affects final classification layers due to representation drift \cite{davari2022probing, ramasesh2020anatomycatastrophicforgettinghidden}. While some class-incremental methods address drift \cite{yu2020semantic, goswami2024resurrecting, magistrielastic}, task-incremental learning (TIL) often relies on fixed task-specific classifiers, failing to adapt them to evolving features. 

To address this classifier adaptation challenge in TIL, we introduce Prototype-Augmented Hypernetworks ($PAH$). ~$PAH$ employs a shared hypernetwork, conditioned on \textit{learnable task prototypes}, to dynamically generate task-specific classifier heads. This approach, combined with knowledge distillation \cite{li2017} to preserve representations, effectively mitigates forgetting without requiring stored classifier heads. Our core contributions include a) an effective prototype-based task embedding mechanism, b) its integration with hypernetworks for dynamic head generation, and c) demonstrating state-of-the-art performance with near-zero forgetting on standard CL benchmarks. 

\section{Related Work}
\label{sec:Related Work}

Continual learning aims to learn sequential tasks without catastrophic forgetting \cite{MCCLOSKEY1989109, FRENCH1999128}. Major approaches include regularization methods that penalize changes to important weights (EWC \cite{doi:10.1073/pnas.1611835114}), rehearsal methods that replay stored or generated data (DER \cite{NEURIPS2020_b704ea2c}, GCR \cite{Tiwari_2022_CVPR}), and architecture-based methods that allocate new parameters per task (PNN \cite{Rusu}). These strategies often involve trade-offs regarding memory usage, computational overhead, or plasticity limits.



Our work intersects hypernetworks and prototype‑based methods for continual learning.  
Hypernetworks \cite{Ha} generate parameters for a target network and have been adapted to CL by conditioning on task identity to produce either full‑network weights or selected layers \cite{vonoswald, hemati2023partial}.  
PAH keeps the same generator idea but feeds it \textit{learnable task prototypes} to dynamically generate weights only for the task-specific classifier head, avoiding the need to store separate heads.
The prototypes--compact class exemplars--have a long history in few‑shot learning \cite{snell} and as anchors for replay or nearest‑neighbour decisions in CL \cite{Rebuffi, isele}.  
In PAH prototypes act instead as a task embedding that conditions the hypernetwork; while knowledge distillation \cite{li2017} limits forgetting.  
The approach differs from Prototype Augmentation and Self‑Supervision (PASS) \cite{zhu2021prototype}, which keeps one prototype per class and feeds those fixed vectors directly to the classifier. $PAH$ stores no classifier heads and never sends prototypes directly to softmax at test time.

\begin{figure*}[t]
    \centering
   
    \includegraphics[width=1\linewidth]{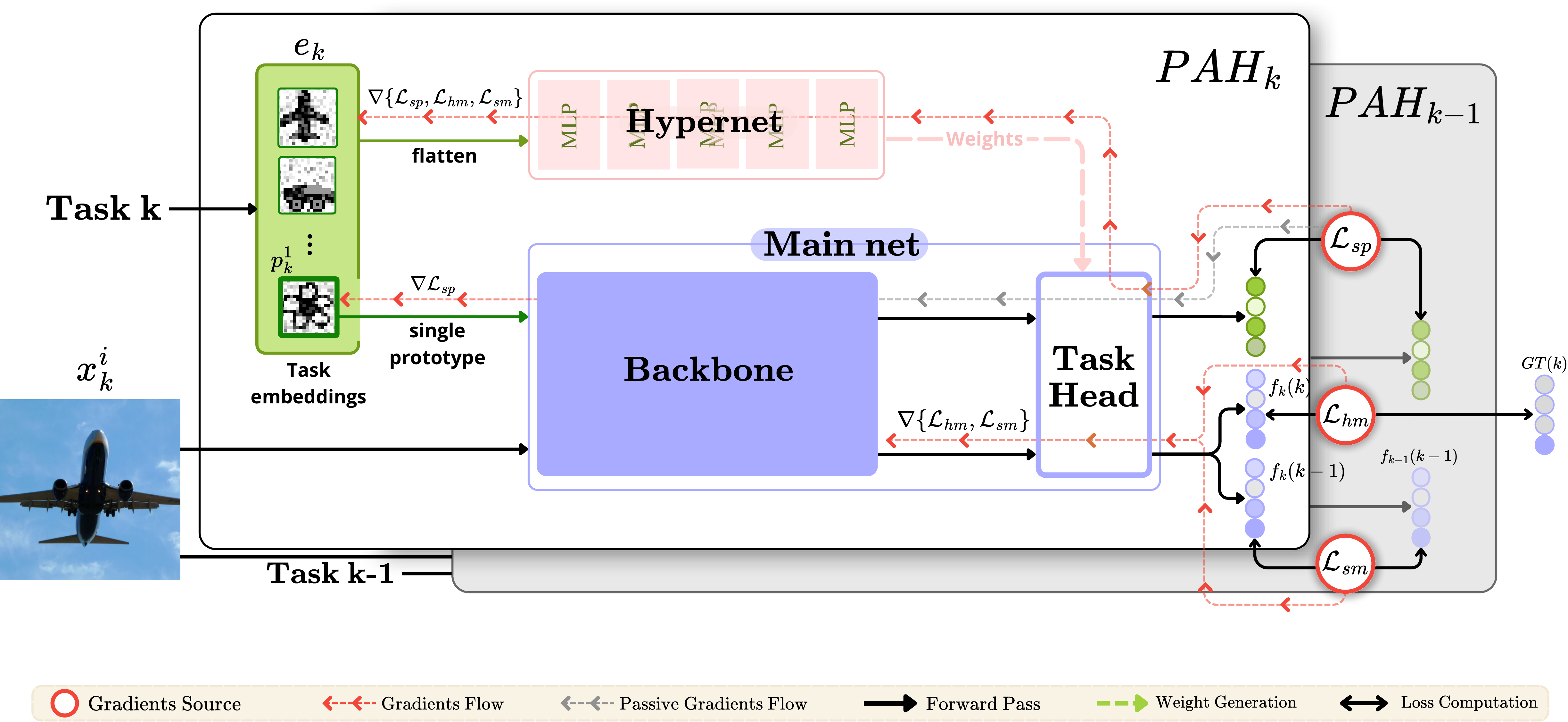}
    \caption{\textbf{Overview of the Prototype-Augmented Hypernetworks ($PAH$) architecture at task k.} The hypernetwork generates task-specific classifier weights conditioned on learnable task prototypes $e_k$, which encode task-specific information. This design eliminates the need to store separate classifier heads for each task and effectively addresses catastrophic forgetting through knowledge distillation.}
    \label{fig:arch}

\vspace{-3mm}
\end{figure*}

\section{Problem Statement}

Formally, we consider a continual multitask learning problem in the task incremental paradigm. The model is presented with a sequence of tasks $\{T_1, T_2, ..., T_K\}$, where each task $T_k$ is associated with a dataset $\mathcal{D}_k = \{(x^i_k, y^i_k)\}_{i=1}^{N_i}$, where $x^i_k$ and $y^i_k$ denote the input data and labels respectively for the $i$-th sample in the $k$-th task. 
Each task consists of a set of classes unique to that task, and the objective is to learn a model that can perform well on all tasks seen so far, without revisiting previous data. Crucially, the task identity is available during training and inference.

\section{Methodology}
\label{sec:methodology}


$PAH$ relies on three main components: (a) a main network, (b) a hypernetwork, and (c) learnable task embeddings.

\paragraph{a) Main Network} The main network comprises a backbone and a task-specific classification head. The backbone extracts features from incoming data, while the classification head assigns class labels based on the current task. Unlike traditional approaches that store a separate head per task, $PAH$ does not require the explicit storage of task-specific head’s weights. Instead, they are generated on demand by the hypernetwork.

Formally, given an input $x_k$ from task $T_k$, the main network processes it as follows:

\begin{equation}
f_{k}(x_k \, | \, k) = \tau (\beta_{\theta} (x_k ) \, | \, \theta_k^\tau)
\end{equation}
where $\beta_{\theta}$ represents the backbone, and $\tau$ is the classification head parameterized by task-specific weights $\theta_k^\tau$.

\paragraph{b) Hypernetwork} A hypernetwork $\Upsilon$ is a net that outputs the weights of another network. In this case, it  takes  a task embedding $e_k$ as input and outputs task-specific classifier weights:
\begin{equation}
\theta_k^\tau = \Upsilon(e_k)
\end{equation}

\paragraph{c) Task Embedding}
The task embedding is the input that the hypernet receives, and it should help the hypernet identify the task and generate the correct weights for it. 
In PAH, it is formed by concatenating learned 2D prototypes (e.g., a compact 10 × 10 feature grid that preserves spatial layout), where each prototype serves as a representative feature for a class within a task. Formally, the task embedding is built as:
\begin{equation}
e_k = \text{Flatten}({p^1_k, p^2_k, ... , p^C_k})
\end{equation}
These prototypes provide the hypernetwork with a compact yet rich encoding, enabling it to generate more precise task-specific classifier weights. Unlike fixed embeddings, prototypes are dynamically optimized during training, learning to capture class-specific feature distributions, and ensuring alignment with evolving backbone representations. This approach allows the hypernet to leverage task- and class-specific information, improving adaptability while reducing the need to store separate classifier heads. 

Unless noted, each prototype starts as a random training image from its class, resized to $10\times10$ and normalized; we call this \textit{semantic initialization}.


\subsection{Loss Functions} \label{subsec:loss_functions}

We use several loss terms tailored to address both classification accuracy and the stability-plasticity trade-off. In addition to the main image-based losses ($L_{hm}$ and $L_{sm}$), we use prototype-based losses to refine task embeddings ($L_{sp}$).

\paragraph{\textbf{Hard Loss Main ($L_{hm}$)}:}
        This term uses the standard cross-entropy loss over the real input images to classify:
        \begin{equation}
            L_{hm} = - \sum_{c=1}^{C} y_{c} \log \hat{y}_{c}            
        \end{equation}
        Here, $y_{c}$ is the ground-truth label for class $c$, and $\hat{y}_{c}$ is the model's predicted probability. 
        Minimizing $L_{hm}$ ensures accurate classification on the current task. As illustrated in Figure \ref{fig:arch}, this loss backpropagates through multiple components of the network, updating the backbone, hypernetwork, and task embeddings collectively.

\paragraph{\textbf{Soft Loss Main ($L_{sm}$)}:}
        This KL-divergence term preserves knowledge from previous tasks. When learning task $T_k$, for an input $x_k$ from $T_k$:
        \begin{equation}
            L_{sm} = \frac{1}{k\text{-}1} \sum_{j=1}^{k-1} \text{KL}(f_{k-1}(x_k \,|\, j) \;\|\; f_{k}(x_k \,|\, j))
            \label{eq:lsm_corrected}
        \end{equation}
        Here, $f_{k-1}(x_k \,|\, j)$ is the output distribution of the model after task $T_{k-1}$ (old model), and $f_{k}(x_k \,|\, j)$ is that of the current (new) model, both evaluated on input $x_k$ using the classifier head generated for a previous task $T_j$.
        Minimizing $L_{sm}$ aligns the new model's predictions for old tasks with those of the old model, reducing catastrophic forgetting and stabilizing the learned representations. This loss updates the backbone, hypernet, and task embeddings.

\paragraph{\textbf{Soft Loss Prototypes ($L_{sp}$)}:}
        This KL-divergence loss adapts prototypes $p_j^c$ from previous tasks $T_j$ (where $j<k$) to the backbone's evolving feature space, assuming a fixed number of $C$ classes per task. For each $p_j^c$:
        \begin{equation}
            L_{sp} = \frac{1}{k\text{-}1} \sum_{j=1}^{k-1} \sum_{c=1}^{C} \text{KL}(f_{k-1}(p_j^c \,|\, j) \;\|\; f_{k}(p_j^c \,|\, j))
            \label{eq:lsp_simplified}
        \end{equation}
        Here, $p_j^c$ is the prototype for class $c$ of task $T_j$. $f_{k-1}(p_j^c \,|\, j)$ is the output of the model after $T_{k-1}$ (old model), and $f_{k}(p_j^c \,|\, j)$ is that of the current model, both using input $p_j^c$ and task $T_j$'s classifier head $\theta_j^\tau$.
        Minimizing $L_{sp}$ refines the prototypes $p_j^c$. Crucially:
        \begin{itemize}
            \item The old model $f_{k-1}$ is frozen.
            \item For $f_{k}$ during the $L_{sp}$ step, the backbone and hypernetwork are also frozen. Gradients from $L_{sp}$ only update the parameters of the prototypes $p_j^c$. This allows prototypes to adapt while the backbone and hypernetwork serve as a stable reference.
        \end{itemize}    

\noindent\textbf{Total Loss:}
We optimize the joint objective
$L_{\text{total}} = L_{hm} + \lambda_{sm} L_{sm} + \lambda_{sp} L_{sp}$,
where the scalars $\lambda_{sm}$ and $\lambda_{sp}$ weight the
distillation and prototype terms, respectively.

\begin{table*}
\caption{\textbf{Performance comparison of $PAH$ against state-of-the-art baselines on Split-CIFAR100 and TinyImageNet.} Models marked with $\dagger$ are evaluated on the 10-task TinyImageNet setting, while the rest use the 20-task setting. Mean $\pm$ std over 3 random seeds.}
\label{tab:pah_results}
    \centering
    \resizebox{0.86\linewidth}{!}{
\begin{tabular}{llcccc}
\toprule
\multirow{2}{*}{Buffer} & \multirow{2}{*}{Methods} & \multicolumn{2}{c}{Split-CIFAR100} & \multicolumn{2}{c}{TinyImageNet} \\ 
\cmidrule(lr){3-4} \cmidrule(lr){5-6}
 &  & $ACCURACY\uparrow$ & $FORGETTING\downarrow$ & $ACCURACY\uparrow$ & $FORGETTING\downarrow$ \\
\cmidrule(lr){1-2} \cmidrule(lr){3-4} \cmidrule(lr){5-6}
200 & GCR~\citep{Tiwari_2022_CVPR}\textsuperscript{$\dagger$}          
& 64.24$\pm$0.83 & 24.12$\pm$1.17 
& 42.11$\pm$1.01 & 40.36$\pm$1.08 \\
200 & DER~\citep{NEURIPS2020_b704ea2c}\textsuperscript{$\dagger$}            
& 63.09$\pm$1.09 & 25.98$\pm$1.55 
& 42.27$\pm$0.9 & 40.43$\pm$1.05 \\
200 & CCLIS~\citep{li2024contrastive}\textsuperscript{$\dagger$}  
& 72.93$\pm$0.46 & 14.17$\pm$0.20
& 48.29$\pm$0.78 & 33.20$\pm$0.75 \\ 
--- & SI~\citep{zenke2017continual}       
& 63.58$\pm$0.37 & 27.98$\pm$0.34
& 44.96$\pm$2.41 & 26.29$\pm$1.40 \\ 
200-250 & A-GEM~\citep{chaudhry}           
& 59.81$\pm$1.07 & 30.08$\pm$0.91
& 60.45$\pm$0.24 & 24.94$\pm$1.24 \\
--- & PNN~\citep{Rusu}
& 66.58$\pm$1.00  & --- 
& 62.15$\pm$1.35  & --- \\ 
200-400 & HN-2~\citep{hemati2023partial}  
& 62.80$\pm$1.60 & 4.10$\pm$0.50 
& 39.90$\pm$0.40 & 10.20$\pm$2.20 \\ 
200-400 & HN-3~\citep{hemati2023partial}  
& 58.80$\pm$1.00 & 7.40$\pm$0.90
& 31.40$\pm$1.80 & 5.20$\pm$0.70 \\ 
\cmidrule(lr){1-2} \cmidrule(lr){3-4} \cmidrule(lr){5-6}
\cmidrule(lr){1-2} \cmidrule(lr){3-4} \cmidrule(lr){5-6}
--- & \textbf{ $PAH$ (Ours)}\textsuperscript{$\dagger$}
& & 
& \textbf{60.48$\pm$0.148} & \textbf{3.24$\pm$0.6} \\
--- & \textbf{ $PAH$ (Ours)}
& \textbf{74.46$\pm$0.08} & \textbf{1.71$\pm$0.02} 
& \textbf{63.65$\pm$1.94} & \textbf{4.43$\pm$0.07} \\
\bottomrule
\end{tabular}
}
\end{table*}

\section{Experiments}

We evaluate our proposed \textit{Prototype-Augmented Hypernetworks} on standard continual learning benchmarks under the task-incremental learning paradigm. All experiments are conducted in the cold start setting, where the dataset is equally divided into a sequence of tasks, and task identities are provided during both training and evaluation. We compare our method to relevant CL baselines and assess its robustness across various datasets, architectures, and ablations. We report Average Accuracy (AA) and Forgetting (FM) as defined by~\cite{chaudhry2018riemannian}.

\subsection{Datasets}

Our evaluation is conducted on two widely-used benchmarks for continual multitask learning: \textbf{1) Split-CIFAR100}: The CIFAR-100 dataset \cite{krizhevsky2009learning} is split into 10 tasks with 10 classes per task. \textbf{2) TinyImageNet}: A reduced version of ImageNet \cite{krizhevsky2012imagenet}, divided into two configurations: 10 tasks with 20 classes each and 20 tasks with 10 classes each.

\subsection{Baselines}

To evaluate the effectiveness of our method, we compare its performance against a diverse set of baselines that represent key paradigms in continual learning, including replay-based methods, regularization strategies, and architecture-specific approaches. Specifically, we use the following baselines: Gradient Coreset Replay (GCR) \cite{Tiwari_2022_CVPR}, Dark Experience Replay (DER) \cite{NEURIPS2020_b704ea2c}, Synaptic Intelligence (SI) \cite{zenke2017continual}, Averaged Gradient Episodic Memory (A-GEM) \cite{chaudhry}, Progressive Neural Networks (PNN) \cite{Rusu}, Partial Hypernetworks (HN-2 and HN-3) \cite{hemati2023partial}, and Contrastive Continual Learning with Importance Sampling (CCLIS) \cite{li2024contrastive}. These baselines were chosen for their known effectiveness and coverage of different ways of addressing catastrophic forgetting.


Replay-based baselines (GCR, DER, CCLIS) are evaluated with the smallest buffer size reported in their papers. This keeps the comparison fair, because $PAH$ does not directly rely on stored samples, and larger buffers would amplify the inherent advantage replay methods gain from storing past samples.


We selected HN-2 and HN-3, which freeze two or three ResNet-18 blocks and generate only the final-layers weights, giving a good accuracy–efficiency trade-off. HN-F is omitted for its heavy computation, leaving HN-2/3 as the practical hypernetwork baselines for $PAH$.

All methods use a ResNet18 backbone. The baseline results were taken directly from their respective sources: PNN and SI from \cite{kumar2024}, GCR from \cite{Tiwari_2022_CVPR}, DER and CCLIS from \cite{li2024contrastive}, A-GEM from \cite{madaan2022representational}, and HN-2/HN-3 from \cite{hemati2023partial}. We compare all methods on standard benchmarks, including Split-CIFAR100 and TinyImageNet.

  
\section{Results}

Table~\ref{tab:pah_results} compares $PAH$ with state-of-the-art baselines on Split-CIFAR100 and TinyImageNet. $PAH$ consistently outperforms competing methods, achieving low forgetting and robust accuracy across tasks.

On Split-CIFAR100, $PAH$ attains 74.46\% accuracy with only 1.71\% forgetting. This surpasses methods like A-GEM~\citep{chaudhry} and SI~\citep{zenke2017continual}, which show both lower accuracy and higher forgetting. While CCLIS~\citep{li2024contrastive} has similar accuracy (72.93\%), its forgetting (14.17\%) is substantially higher, highlighting $PAH$’s strength in knowledge retention.

On TinyImageNet, $PAH$ achieves 63.65\% accuracy and just 4.43\% forgetting. In contrast, replay-based methods such as GCR~\citep{Tiwari_2022_CVPR} and DER~\citep{NEURIPS2020_b704ea2c} suffer from around 40\% forgetting despite storing samples. Similarly, hypernetwork-based methods HN-2 and HN-3~\citep{hemati2023partial} lag behind $PAH$ in both metrics. These results show the effectiveness of combining hypernets with prototypes to achieve minimal forgetting and strong accuracy, establishing $PAH$ as a new benchmark in CL. This low classifier-level forgetting suggests PAH effectively mitigates the representation drift impact where it is most damaging.

\subsection{Ablation Studies}
\label{subsec:ablation}

\definecolor{baselinecolor}{gray}{.9} 
\newcommand{\baselineconfig}[1]{\cellcolor{baselinecolor}{#1}}

\begin{table}[t]
\centering
\caption{\textbf{Ablation Study:} Impact of various factors on \textit{PAH}. All experiments use Split-CIFAR100. Default settings marked in \colorbox{baselinecolor}{gray}.}
\label{tab:pah2di_ablation_combined}

\subfigure[
\centering
\textbf{ Prototype Shape}
\label{tab:proto_shape}
]{
\begin{minipage}{0.29\linewidth}
\centering
\footnotesize
\setlength{\tabcolsep}{4pt}
\renewcommand{\arraystretch}{1.06}
\begin{tabular}{lcc}
\shline
Shape & \textbf{AA}$\uparrow$ & \textbf{FM}$\downarrow$\\
\shline
$5\times5$    & 73.72 & 1.89 \\
\baselineconfig{$10\times10$} & \baselineconfig{\textbf{74.46}} & \baselineconfig{\textbf{1.71}}\\
$16\times16$  & 73.52 & 1.76 \\
$20\times20$  & 71.95 & 2.66 \\
$30\times30$  & 63.87 & 8.41 \\
\shline
\end{tabular}
\end{minipage}
}
\hfill
\subfigure[
\centering
\textbf{ Stability}
\label{tab:stability}
]{
\begin{minipage}{0.5\linewidth}
\centering
\setlength{\tabcolsep}{4pt}
\renewcommand{\arraystretch}{0.7}
\begin{tabular}{lcc}
\shline
Stability & \textbf{AA}$\uparrow$ & \textbf{FM}$\downarrow$\\
\shline
0.1  & 69.27 & 7.32 \\
0.25 & 72.52 & 2.94 \\
\baselineconfig{0.5}  & \baselineconfig{\textbf{74.46}} & \baselineconfig{1.71}\\
1.0  & 73.48 & 0.83 \\
1.5  & 72.89 & \textbf{0.69}\\
\shline
\end{tabular}
\end{minipage}
}\\[6pt]


\subfigure[
\centering
\textbf{ $L_{sp}$ Weight}
\label{tab:lsp_weight}
]{
\begin{minipage}{0.29\linewidth}
\centering
\footnotesize
\setlength{\tabcolsep}{4pt}
\renewcommand{\arraystretch}{0.8}
\begin{tabular}{lcc}
\shline
$L_{sp}$ weight & \textbf{AA}$\uparrow$ & \textbf{FM}$\downarrow$\\
\shline
0.0 & 72.93 & 1.56 \\
0.5 & 74.32 & \textbf{0.97} \\
\baselineconfig{1.0} & \baselineconfig{\textbf{74.46}} & \baselineconfig{1.71}\\
2.0 & 74.08 & 1.24 \\
\shline
\end{tabular}
\end{minipage}
}
\hfill
\subfigure[
\centering
\textbf{ Prototype Init.}
\label{tab:proto_init}
]{
\begin{minipage}{0.50\linewidth}
\centering
\footnotesize
\setlength{\tabcolsep}{4pt}
\renewcommand{\arraystretch}{1.06}
\begin{tabular}{lcc}
\shline
Init & \textbf{AA}$\uparrow$ & \textbf{FM}$\downarrow$\\
\shline
Random  & 72.54 & 2.32 \\
\baselineconfig{Semantic} & \baselineconfig{\textbf{74.46}} & \baselineconfig{\textbf{1.71}}\\
\shline
\end{tabular}
\end{minipage}
}
\vspace{-0.5cm}

\end{table}

Ablation studies on Split-CIFAR100 validated key components of $PAH$ (see Table~\ref{tab:pah2di_ablation_combined}). Tuning the overall \textbf{stability} coefficient (controlling the influence of distillation losses $L_{sm}$ and $L_{sp}$) proved crucial; a value of 0.5 yielded the best trade-off between accuracy (74.46\%) and forgetting (1.71\%), avoiding excessive forgetting seen at lower values and the drop in plasticity at higher values. The \textbf{$L_{sp}$ weight}, which specifically aligns prototypes, is vital: performance significantly dropped when its weight was 0, while a weight of 1.0 provided the optimal balance, demonstrating its importance for maintaining prototype relevance. \textbf{Prototype Shape} also impacted performance, $10 \times 10$ prototypes being the best (74.46\% AA, 1.71\% FM); smaller shapes ($5 \times 5$) were less expressive, while larger ones ($16 \times 16$ and above) showed diminishing returns or signs of overfitting. Critically, \textbf{prototype initialization} using semantic information (starting from real class images) significantly outperformed random initialization (74.46\% vs 72.54\% AA, 1.71\% vs 2.32\% FM), providing a better starting point for optimization, leading to faster convergence and improved stability. These studies confirm that carefully configured learnable prototypes, guided by appropriate distillation, are central to PAH's effectiveness.


\section{Conclusion}

Catastrophic forgetting in classifier layers remains a key challenge in CL. We introduced Prototype-Augmented Hypernetworks, a novel TIL approach where a hypernetwork dynamically generates classifier heads conditioned on learnable task prototypes. Combined with knowledge distillation, $PAH$ prevents forgetting without storing task-specific heads or large replay buffers. Experiments demonstrate that $PAH$ achieves SOTA performance on standard benchmarks like Split-CIFAR100 and TinyImageNet, significantly reducing forgetting compared to existing methods. Key to PAH's success is the use of semantically initialized, learnable prototypes that adapt alongside the feature extractor, providing stable and informative conditioning for the hypernetwork. 

\section*{Acknowledgments}
This work was supported by the following Grant Numbers: PID2020120311RB-I00 and RED2022–134964-T and funded by MCINAEI/10.13039/501100011033. 

{
    \small
    \bibliographystyle{ieeenat_fullname}
    \bibliography{main}
}


\end{document}

%% file: sec/0_abstract.tex
\begin{abstract}
Continual learning (CL) aims to learn a sequence of tasks  without forgetting prior knowledge, but gradient updates for a new task often overwrite the weights learned earlier, causing catastrophic forgetting (CF). We propose \textit{Prototype-Augmented Hypernetworks} ($PAH$), a framework where a single hypernetwork, conditioned on learnable task prototypes, dynamically generates task-specific classifier heads on demand.  To mitigate forgetting, $PAH$ combines cross-entropy with dual distillation losses, one to align logits and another to align prototypes, ensuring stable feature representations across tasks. Evaluations on Split-CIFAR100 and TinyImageNet demonstrate that PAH achieves state-of-the-art performance, reaching 74.5\% and 63.7\% accuracy with only 1.7\% and 4.4\% forgetting, respectively, surpassing prior methods without storing samples or heads.\vspace{0.01cm}
\end{abstract}